\documentclass[sigconf]{acmart}
\setcopyright{none}  
\renewcommand\footnotetextcopyrightpermission[1]{}
\AtBeginDocument{%
  }

\acmConference[JCRAI 2024]{Make sure to enter the correct
  conference title from your rights confirmation email}{September 13-15,
  2024}{Shanghai, China}

\usepackage{multicol}
\usepackage{amsmath}
\usepackage{graphicx}
\graphicspath{{Images/}}
\usepackage{subfigure}
\usepackage{color}
\usepackage{stfloats}
\usepackage{hyperref}
\usepackage{fancyhdr}

\begin{document}

\title{FeaKM: Robust Collaborative Perception under Noisy Pose Conditions}


\author{Jiuwu Hao}
\orcid{0009-0009-2002-7984}
\affiliation{%
  \institution{School of Artificial Intelligence, University of Chinese Academy of Sciences}
  \institution{Institute of Automation of the Chinese Academy of Sciences}
  \streetaddress{95 Zhongguancun East Road}
  \city{Beijing}
  \country{China}
  \postcode{100190}
}
\email{haojiuwu2022@ia.ac.cn}

\author{Liguo Sun}
\affiliation{%
  \institution{Institute of Automation of the Chinese Academy of Sciences}
  \city{Beijing}
  \country{China}}
\email{liguo.sun@ia.ac.cn}

\author{Ti Xiang}
\affiliation{%
  \institution{School of Artificial Intelligence, University of Chinese Academy of Sciences}
  \institution{Institute of Automation of the Chinese Academy of Sciences}
  \city{Beijing}
  \country{China}}
\email{xiangti@mails.ucas.ac.cn}

\author{Yuting Wan}
\affiliation{%
  \institution{School of Artificial Intelligence, University of Chinese Academy of Sciences}
  \institution{Institute of Automation of the Chinese Academy of Sciences}
  \city{Beijing}
  \country{China}}
\email{wanyuting2022@ia.ac.cn}

\author{Haolin Song}
\affiliation{%
  \institution{School of Artificial Intelligence, University of Chinese Academy of Sciences}
  \institution{Institute of Automation of the Chinese Academy of Sciences}
  \city{Beijing}
  \country{China}}
\email{songhaolin2024@ia.ac.cn}

\author{Pin Lv}
\authornote{Corresponding author}
\affiliation{%
  \institution{Institute of Automation of the Chinese Academy of Sciences}
  \city{Beijing}
  \country{China}}
\email{pin.lv@ia.ac.cn}

\renewcommand{\shortauthors}{Jiuwu Hao et al.}

\begin{abstract}
    Collaborative perception is essential for networks of agents with limited sensing capabilities, enabling them to work together by exchanging information to achieve a robust and comprehensive understanding of their environment. However, localization inaccuracies often lead to significant spatial message displacement, which undermines the effectiveness of these collaborative efforts. To tackle this challenge, we introduce \textbf{FeaKM}, a novel method that employs \textbf{Fea}ture-level \textbf{K}eypoints \textbf{M}atching to effectively correct pose discrepancies among collaborating agents. Our approach begins by utilizing a confidence map to identify and extract salient points from intermediate feature representations, allowing for the computation of their descriptors. This step ensures that the system can focus on the most relevant information, enhancing the matching process. We then implement a target-matching strategy that generates an assignment matrix, correlating the keypoints identified by different agents. This is critical for establishing accurate correspondences, which are essential for effective collaboration. Finally, we employ a fine-grained transformation matrix to synchronize the features of all agents and ascertain their relative statuses, ensuring coherent communication among them. Our experimental results demonstrate that FeaKM significantly outperforms existing methods on the DAIR-V2X dataset, confirming its robustness even under severe noise conditions. The code and implementation details are available at \href{https://github.com/uestchjw/FeaKM}{\textcolor{purple}{https://github.com/uestchjw/FeaKM}}.
\end{abstract}



\maketitle

\section{Introduction}
    Autonomous driving has become a hot-spot issue both in academia and industry due to the increasing demand for smart transportation. In recent years, single-vehicle perception has been rapidly improved with advancements of deep learning based methods \cite{BEVFusion, BEVFormer, BEVFormer_v2}. However, individual perception suffers from inherent limitations, such as sparse remote data and occlusion, resulting in incomplete and unsatisfying perception result. Thus, several efforts \cite{CoCa3D, V2VNet, v2xvit} have been made to explore collaborative perception methods, which leverages agent-to-agent communication to share perception information. 
    \par
    Since the roadside infrastructure, which is often installed in key scenes, can always be a environmental information provider, vehicle-to-infrastructure (V2I) perception system has great potential to revolutionize the self-driving industry. However, due to sensor aging and the impact of some uncontrollable factors, it is vital to determine the status of the infrastructure and ensure that correct information is provided to the vehicle. Most existing methods \cite{CoAlign, FeaCo} rely on the initial values of the given pose, making it impossible to perform pose rectification and verify sensor status in cases of large errors. So here is the question: \textit{How to verify the status of infrastructure and utilize its data for collaboration in situations where there may be significant errors?}
    \par
    Motivated by this limitation, we propose a novel collaboration framework FeaKM, which could deal with arbitrary pose errors (even huge errors). The core idea is to leverage keypoints matching to obtain the relative position between agents. Once an object can be observed by both the vehicle and infrastructure, its spatial position and neighboring descriptors are consistent in the two feature maps. Since FeaKM does not rely on the initial values of the pose, this method can determine the state of the infrastructure by comparing the given pose with the calculated pose. Then, FeaKM will use the calculated pose for collaboration if there is a significant difference between the two poses.
    \par
    In summary, the main contributions of this work are:
    \begin{itemize}
    \item[$\bullet$]We propose FeaKM, a novel collaborative framework, which can maintain perception precision and robustness in various pose error conditions.
    \item[$\bullet$]A pose-error rectification module based on keypoints matching is designed to determine infrastructure's status and to obtain the relative positions.
    \item[$\bullet$]Experiments show that our FeaKM can achieve more accurate collaborative perception results in various noisy conditions.
    \end{itemize}

\begin{figure*}[ht]
    \centering
    \includegraphics[scale=0.35]{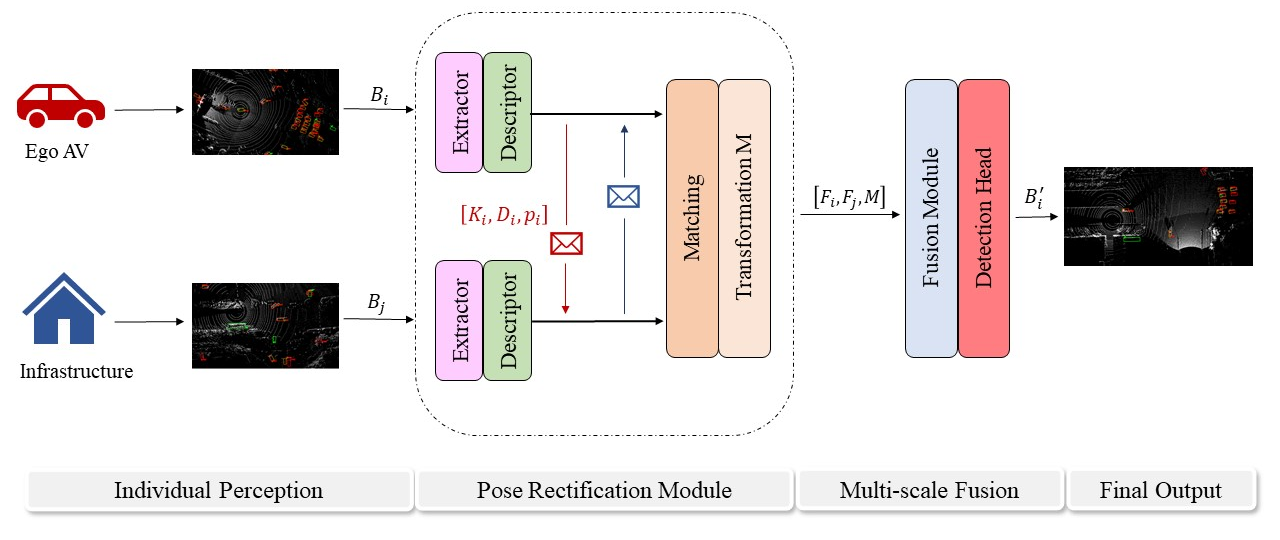}
    \caption{Overall architecture of proposed FeaKM.}
    \label{fig:1}
\end{figure*}

\section{RELATED WORKS}
\subsection{Collaborative Perception}
    Collaborative perception has raised great interest recently for its potential in autonomous driving tasks. With the emergence of several high-quality datasets, including V2X-Sim \cite{V2X-Sim}, OPV2V \cite{OPV2V} and DAIR-V2X \cite{DAIR-V2X}, etc., a large body of research has been devoted to improving the effectiveness and robustness of multi-agent collaborative perception \cite{CoCa3D,  Where2comm, CORE, HEAL}. For example, Where2comm \cite{Where2comm} proposes a spatial confidence map to pursue a trade-off between communication bandwidth and perception performance. Based on a learning-to-reconstruction formulation, CORE \cite{CORE} presents a conceptually simple but effective model which can offer reasonable and clear supervision to inspire more effective collaboration and eventually promote perception tasks. To address the open heterogeneous problem, HEAL \cite{HEAL} establishes a unified feature space and aligns new agents with an innovative backward alignment. However, these models assume that the vehicle localization is always accurate, which is difficult to achieve in realistic scenarios.

\subsection{Pose Rectification}
    Localization error is a serious challenge encountered by collaborative perception in practical applications, leading to spatial misalignment of features during fusion process. Inaccurate poses caused by sensor accuracy or unknown interference would result in worse collective perception performance. Therefore, many methods attempt to design pose-error rectification module or robust network to alleviate error effects. FPV-RCNN \cite{fpvrcnn} classifies the selected keypoints and use maximum consensus algorithm to find correspondences between agents. In order to handle pose errors, V2X-ViT \cite{v2xvit} present a robust cooperative framework, which consists of multi-agent self-attention and multi-scale window attention. CoAlign \cite{CoAlign} proposes to use agent-object graph optimazation to enhance pose consistency and rectify localization errors. Based on feature-level consensus matching, FeaCo \cite{FeaCo} also devises a pose-error rectification module PRM to align features. In this work, we propose an efficient framework FeaKM to deal with more serious pose inaccuracies.

    \begin{table*}
  \caption{Detection performance AP@0.5 on DAIR-V2X with pose noises.}
  \label{table1}
  \begin{tabular}{c|ccccccccccc}
    \toprule
    Method/Noise & 0.0/0.0 & 0.0/0.2 & 0.4/0.4 & 0.6/0.6 & 0.8/0.8 & 1.0/1.0 & 1.2/1.2 & 1.4/1.4 & 1.6/1.6 & 1.8/1.8 & 2.0/2.0 \\
    \midrule
    MASH \cite{MASH} & 0.400 & 0.400 & 0.400 & 0.400 & 0.400 & 0.400 & 0.400 & 0.400 & 0.400 & 0.400 & 0.400 \\
    F-Cooper \cite{F-Cooper}& 0.740 & 0.723 & 0.703 & 0.693 & 0.676 & 0.672 & 0.670 & 0.662 & 0.659 & 0.658 & 0.650 \\
    V2X-ViT \cite{v2xvit} & 0.713 & 0.706 & 0.693 & 0.680 & 0.670 & 0.664 & 0.654 & 0.651 & 0.643 & 0.638 & 0.637 \\
    CoAlign \cite{CoAlign} & 0.762 & 0.746 & \textbf{0.725} & 0.701 & 0.687 & 0.680 & 0.671 & 0.664 & 0.659 & 0.658 & 0.654 \\
    Ours(4pairs)   & 0.726 & 0.718 & 0.711 & 0.703 & \textbf{0.697} & \textbf{0.697} & \textbf{0.695} & \textbf{0.694} & \textbf{0.694} & \textbf{0.694} & \textbf{0.692} \\
    Ours(8pairs) & \textbf{0.768} & \textbf{0.753} & 0.716 & \textbf{0.705} & 0.690 & 0.686 & 0.678 & 0.673 & 0.670 & 0.667 & 0.668 \\
    \bottomrule
  \end{tabular}
\end{table*}

\begin{table*}
  \caption{Detection performance AP@0.7 on DAIR-V2X with pose noises.}
  \label{table2}
  \begin{tabular}{c|ccccccccccc}
    \toprule
    Method/Noise & 0.0/0.0 & 0.0/0.2 & 0.4/0.4 & 0.6/0.6 & 0.8/0.8 & 1.0/1.0 & 1.2/1.2 & 1.4/1.4 & 1.6/1.6 & 1.8/1.8 & 2.0/2.0 \\
    \midrule
    MASH \cite{MASH} & 0.244 & 0.244 & 0.244 & 0.244 & 0.244 & 0.244 & 0.244 & 0.244 & 0.244 & 0.244 & 0.244 \\
    F-Cooper \cite{F-Cooper}& 0.579 & 0.572 & 0.563 & 0.555 & 0.553 & 0.546 & 0.546 & 0.542 & 0.540 & 0.534 & 0.533 \\
    V2X-ViT \cite{v2xvit} & 0.544 & 0.539 & 0.536 & 0.531 & 0.526 & 0.523 & 0.522 & 0.520 & 0.516 & 0.512 & 0.510 \\
    CoAlign \cite{CoAlign} & 0.625 & 0.596 & \textbf{0.582} & \textbf{0.575} & 0.570 & 0.567 & 0.564 & 0.559 & 0.558 & 0.557 & 0.555 \\
    Ours(4pairs) & 0.593 & 0.579 & 0.577 & 0.573 & \textbf{0.572} & \textbf{0.572} & \textbf{0.571} & \textbf{0.571} & \textbf{0.571} & \textbf{0.571} & \textbf{0.571} \\
    Ours(8pairs) & \textbf{0.631} & \textbf{0.601} & 0.577 & 0.573 & 0.570 & 0.566 & 0.561 & 0.563 & 0.560 & 0.558 & 0.558 \\
    \bottomrule
  \end{tabular}
\end{table*}
\section{PROBLEM STATEMENT}
    Consider $N$ agents in the scene. Each agent has the ability to independently perceive the surrounding environment and output detection results. Let $S_i$ and $B_i$ be the observed data and perception output of the $i$th agent, respectively. There are three common collaborative types: early collaboration, intermediate collaboration and late collaboration. Here, we utilize intermediate collaboration, which transmits intermediate features between the cooperative agents. For the $i$th agent, a standard individual perception process works as follows:
    \begin{subequations}
    \begin{align}
        \mathbf{F}_i&=\varPhi _{encoder}\left( \mathbf{S}_i \right) \label{XXa}\\
        \mathbf{I}_{j\rightarrow i}&=\varPhi _{project}\left( \mathbf{F}_j,\left( p_i,p_j \right) \right) \label{XXb}\\
        \mathbf{F}_{i}^{\prime}&=\varPhi _{fusion}\left( \mathbf{F}_i,\left\{ \mathbf{I}_{j\rightarrow i} \right\} _{j=1,2,\cdots ,N} \right) \label{XXc}\\
        \mathbf{B}_i&=\varPhi _{decoder}\left( \mathbf{F}_{i}^{\prime} \right) \label{XXd}
    \end{align}
    \end{subequations}

    where $\mathbf{F}_i$ represent the feature extracted from the $i$th agent, $p_i$ is the 6 degrees of freedom(6DoF) poses with $x_i,y_i,z_i$ the location information and $\theta _i,\phi _i,\psi _i$  the angle information. $
    \mathbf{I}_{j\rightarrow i}$ is the information sent from the $j$th agent to the $i$th agent, and the ego agent will project other agents' features into ego's view according to the relative positions. $\mathbf{F}_{i}^{\prime}$ is the fused feature of the $i$th agent after aggregating other agents' messages.
    \par
    However, the 6DoF pose $p_i$ provided by each agent's positioning module is not always accurate, leading to misalignment and inconsistency during the fusion step \eqref{XXc}. Though it may be affected by noise interference, collaborative perception still has advantages over individual perception. We aim to effectively utilize the information provied by the collaborators even in the presence of severe noise.

\section{Our method}
\subsection{Overall Architecture}
    Now we propose \textbf{FeaKM}, a robust collaboration framework, which utilizes \textbf{Fea}ture-level \textbf{K}eypoints \textbf{M}atching to deal with the misalignment during fusion process, with the architecture shown in Figure \ref{fig:1}. Mathematically, for the $i$th agent, FeaKM works as:
    \begin{subequations}
    \begin{align}
        \mathbf{F}_i&=\varPhi _{encoder}\left( \mathbf{S}_i \right) \label{YYa}\\
        \mathbf{B}_i&=\varPhi _{decoder}\left( \mathbf{F}_i \right) \label{YYb}\\
        \mathbf{T}_{ij}&=\varPhi _{correction}\left( p_i,p_j,\mathbf{B}_i,\mathbf{B}_j \right) \label{YYc}\\
        \mathbf{I}_{j\rightarrow i}&=\varPhi _{project}\left( \mathbf{F}_j,\mathbf{T}_{ij} \right) \label{YYd}\\
        \mathbf{F}_{i}^{\prime}&=\varPhi _{fusion}\left( \mathbf{F}_i,\left\{ \mathbf{I}_{j\rightarrow i} \right\} _{j=1,2,\cdots ,N} \right) \label{YYe}\\
        \mathbf{B}_{i}^{\prime}&=\varPhi _{decoder}\left( \mathbf{F}_{i}^{\prime} \right) \label{YYf}
    \end{align}
    \end{subequations}
    where $\mathbf{B}_i$ is the individual detection output, and $\mathbf{T}_{ij}$ is the transformation matrix from $j$th agent to $i$th agent after the rectification process.
    First each agent obtains perception results based on their own view, and then the pose rectification module uses the detection outputs and initial pose of two agent to derive the fine-grained transformation matrix. Step \eqref{YYd} aligns other agents' intermediate feature to the ego agent based on the fine-grained transformation matrix. In order to further alleviate the effect of localization error, Step \eqref{YYe} adopts a multi-scale fusion method to update the feature. Finally, Step \eqref{YYf} use the same detection head to obtain final results.

\subsection{Pose Rectification Module}
    Here we elaborate Step \eqref{YYc}. The core idea is to find identical objects in features from different agents. Consequently, inspired by \cite{LuShijie}, we utilize a keypoint matching approach to pursue consistency. It consists of two parts, detecting part and matching part, which will be explained in the following subsections. The detecting part selects the interest point and calculates the descriptor of each point, and the matching part obtains the assignment matrix based on the distance similarity and generates the final transformation matrix.
    \subsubsection{Interest Point Extraction} Before communicating with other agents, we need to extract objects of interest from the current view. Considering the limitations of communication bandwidth, we employ spatial confidence map \cite{Where2comm} as the representation of "objectness" level that contains interesting points, which is generated by the detection head $\varPhi _{decoder}$.
    \par
    Specially, the areas with high confidence values indicate there may have objects, while low values often represent background area. We set a detection threshold $\delta$ to determine which pixels are points of interest. The binary spatial map is
    \begin{equation}
        \mathbf{C}_i=\varPhi _{generator}\left( \mathbf{B}_i \right) \in \left[ 0,1 \right] ^{H\times W}
    \end{equation}
    After extracting the keypoints, we use a method similar to SuperPoint \cite{SuperPoint} to obtain the descriptors, which computes $\mathbf{F}_i\in \mathbb{R} ^{H\times W\times D}$ and outputs $\mathbf{D}_i\in \mathbb{R} ^{H\times W\times D}$. In order to save training memory and improve the inference speed, we adopt semi-dense descriptors, following by bicubic interpolation and L2-normalization. 
    \par
    Both extracting keypoints and calculating descriptors are conducted locally, and only the keypoints $\mathbf{K}_i\in \mathbb{R} ^{N\times 2}$, descriptors $\mathbf{D}_i\in \mathbb{R} ^{N\times D}$ and pose will be sent to the collaborators.

    \subsubsection{Target Matching}
    Suppose $i$th agent and $j$th agent have $M$ and $N$ interesting points, indexed by $A:=\left\{ 1,\cdots ,M \right\}$ and $B:=\left\{ 1,\cdots ,N \right\}$, respectively. In order to fully learn the correlation between points, we stacked self- and cross-attention to simultaneously learn the feature representations of two sets. The insight is that if a point can see all the points in its own set and the set to be matched, it is easier to find the feature space which divides the matching relationship through multiple iterations.
    \par
    For each point $i$, we first calculate the query and key vectors $q_i$ and $k_i$ via linear layers. Then we obtain the attention score between points $i$ and $j$ as
    \begin{equation}
        p_{ij}=\mathrm{softmax}_j\left( \mathbf{q}_{i}^{\mathrm{T}}\mathbf{k}_j \right) 
    \end{equation}
    \par
    After aggregation via attention mechanism, the output tensor $z_i$ is as follows:
    \begin{equation}
        \mathbf{z}_i=\sum{p_{ij}\mathbf{v}_j}
    \end{equation}
    where $\mathbf{v}_i$ is the value vector.
    \par
    We use rotary encoding \cite{positional_encoding} to indicate location information of each interesting point, which is critical to capture the relative position of points. After obtaining the semantic representation of each points, we construct the assignment matrix by calculating similarity. Matching pairs vary for different driving scenarios, and we set different minimum required matching pairs. Then We employ the singular value decomposition (SVD) to obtain the final transformation matrix, as shown in Eq \eqref{SVD}.
    \begin{equation}
        \min \left\| UDV^{\mathrm{T}}x-b \right\| ^2
        \label{SVD}
    \end{equation}
\subsection{Multiscale Feature Fusion}
    Though the pose rectification module could obtain a more accurate position, the localization error may still exist, leading to unexpected misalignment during fusion process. To further alleviate this issue, we utilize a multiscale fusion strategy, which fuses features at different spatial scales. The advantage is that the vehicle can receive geometric information from the finer layer and become less sensitive to the pose error due to the coarser layer.
    \par
    Mathematically, the multiscale structure works as
    \begin{subequations}
    \begin{align} 
        \mathbf{F}_{i}^{\left( l \right)}&=\varPhi _{downsample}\left( \mathbf{F}_i \right) \\
        \mathbf{F}_{i}^{^{\prime}\left( l \right)}&=\varPhi _{fuse}\left( \mathbf{F}_{i}^{\left( l \right)},\mathbf{F}_{j}^{\left( l \right)} \right) \\
        \mathbf{F}_{i}^{\prime}&=Cat\left( \left[ \mathbf{F}_{i}^{^{\prime}\left( 1 \right)},\cdots ,\varPhi _{upsample}\left( \mathbf{F}_{i}^{^{\prime}\left( L \right)} \right) \right] \right) 
    \end{align}
    \end{subequations}

\section{experiment}
    We validate the effectiveness and robustness of FeaKM on LiDAR-based 3D object detection task.
\subsection{Dataset and Experiment Settings}
    DAIR-V2X \cite{DAIR-V2X} is a large-scale real-world vehicle-infrastructure cooperative dataset. We set the detection range to $x\in \left[ -100m,100m \right] ,y\in \left[ -40m,40m \right]$ and communication range to 100m. To simulate pose errors, we use Gaussian noise $ \mathcal{N} \left( 0,\sigma _t \right) $ on $ x,y $ and $ \mathcal{N} \left( 0,\sigma _r \right) $ on $ \theta $. We use Adam with learning rate 0.002, batchsize 4 and epoch number 50. We execute the experiment on an NVIDIA RTX 3090 GPU using PyTorch.

\begin{figure}[htbp]
    \centering
    \includegraphics[scale=0.33]{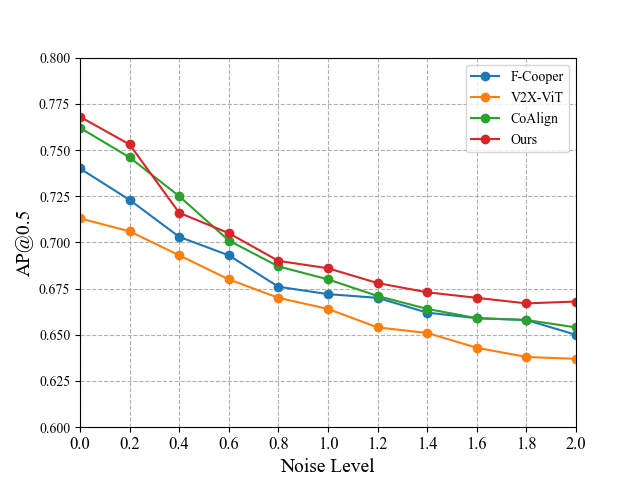}
    \caption{Detection results on DAIR-V2X dataset with different noise errors. Our proposed method outperforms other methods.}
    \label{fig:2}
\end{figure}

\subsection{Quantitative Results}
    We compare the proposed FeaKM with a series of previous methods; see Figure \ref{fig:2}. Table \ref{table1} and \ref{table2} show the average precision (AP) in DAIR-V2X dataset at Intersection-over-Union (IoU) threshold of 0.5 and 0.7. The results show that FeaKM outperforms the previous methods in most noise levels and the improvement is bigger when the noise level is higher. Our FeaKM retains similar performance and demonstrates high robustness under different noisy conditions.
    \par
    As more objects overlap in the scene, the performance of FeaKM improves. As the noise increases, the given pose deviates significantly from the ground truth, resulting in a decrease in the number of matching points. Regardless of the noise settings, FeaKM outperforms all other models.

\begin{table}
  \caption{Ablation studies on DAIR-V2X dataset.}
  \label{tab: ablation}
  \begin{tabular}{cccccl}
    \toprule
    Confidence Map&Multiscale&0.0/0.0&0.4/0.4&0.8/0.8&1.2/1.2\\
    \midrule
       & & 0.741 & 0.704 & 0.682 & 0.678 \\
       & \checkmark & \textbf{0.780} & 0.716 & 0.682 & 0.666 \\
    \checkmark &  & 0.778 & 0.716 & 0.674 & 0.660 \\
    \checkmark & \checkmark & 0.768 & \textbf{0.716} & \textbf{0.690} & \textbf{0.678}\\
  \bottomrule
\end{tabular}
\end{table}

\begin{figure}
    \centering
    \subfigure[F-Cooper]{
    \begin{minipage}[b]{.4\linewidth}
        \centering
        \includegraphics[scale=0.08]{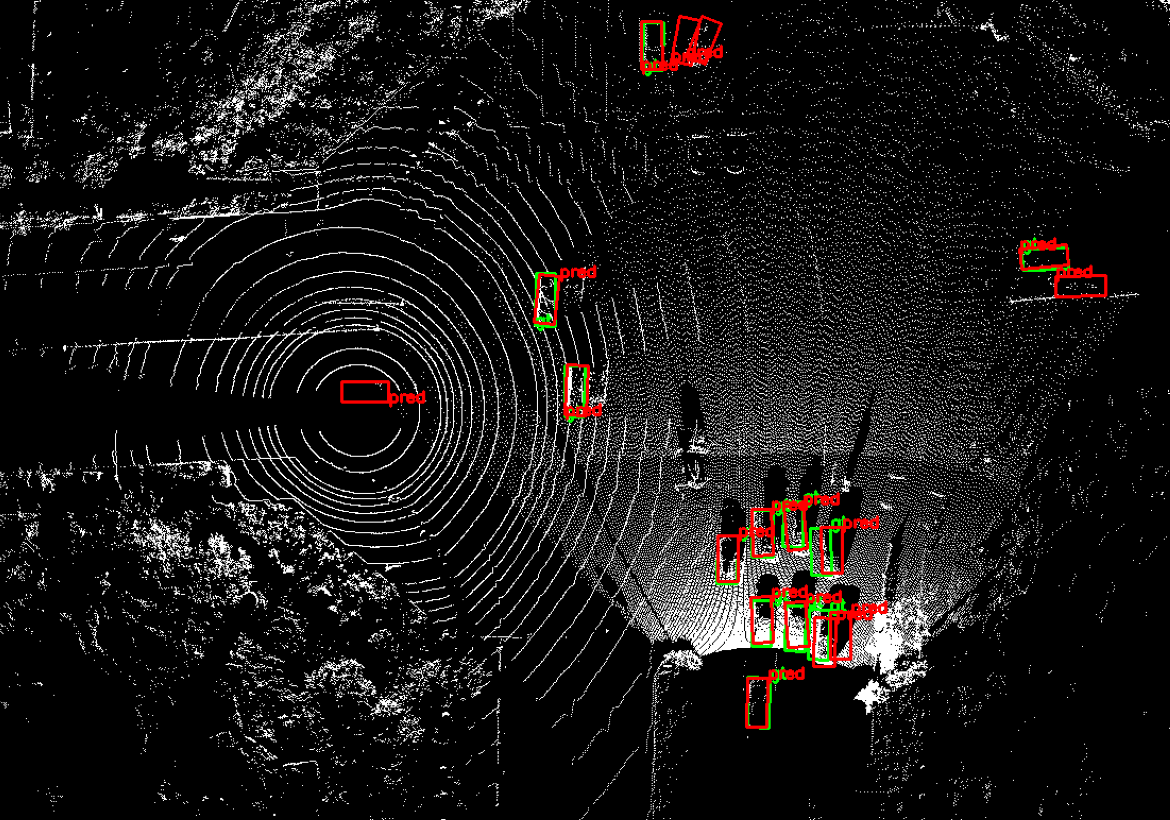}
    \end{minipage}
    }
    \subfigure[V2X-ViT]{
    \begin{minipage}[b]{.4\linewidth}
        \centering
        \includegraphics[scale=0.08]{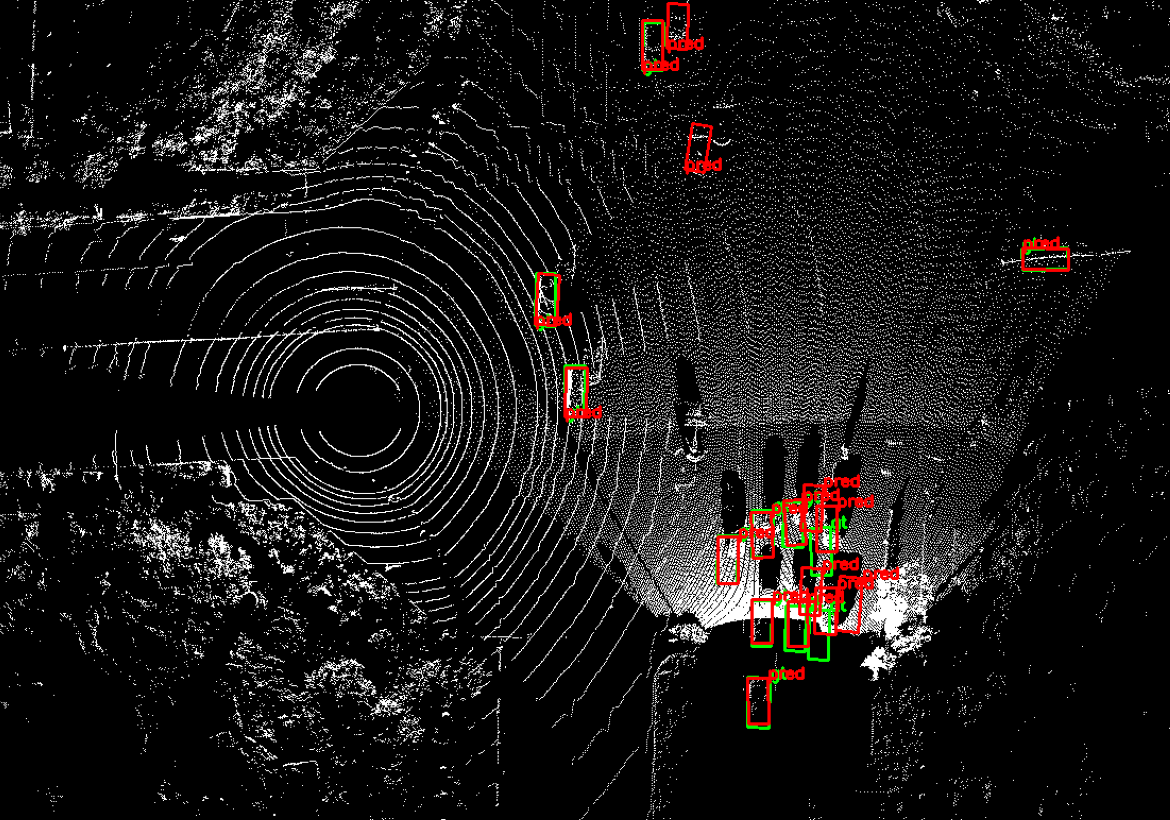}
    \end{minipage}
    }
    \subfigure[CoAlign]{
    \begin{minipage}[b]{.4\linewidth}
        \centering
        \includegraphics[scale=0.08]{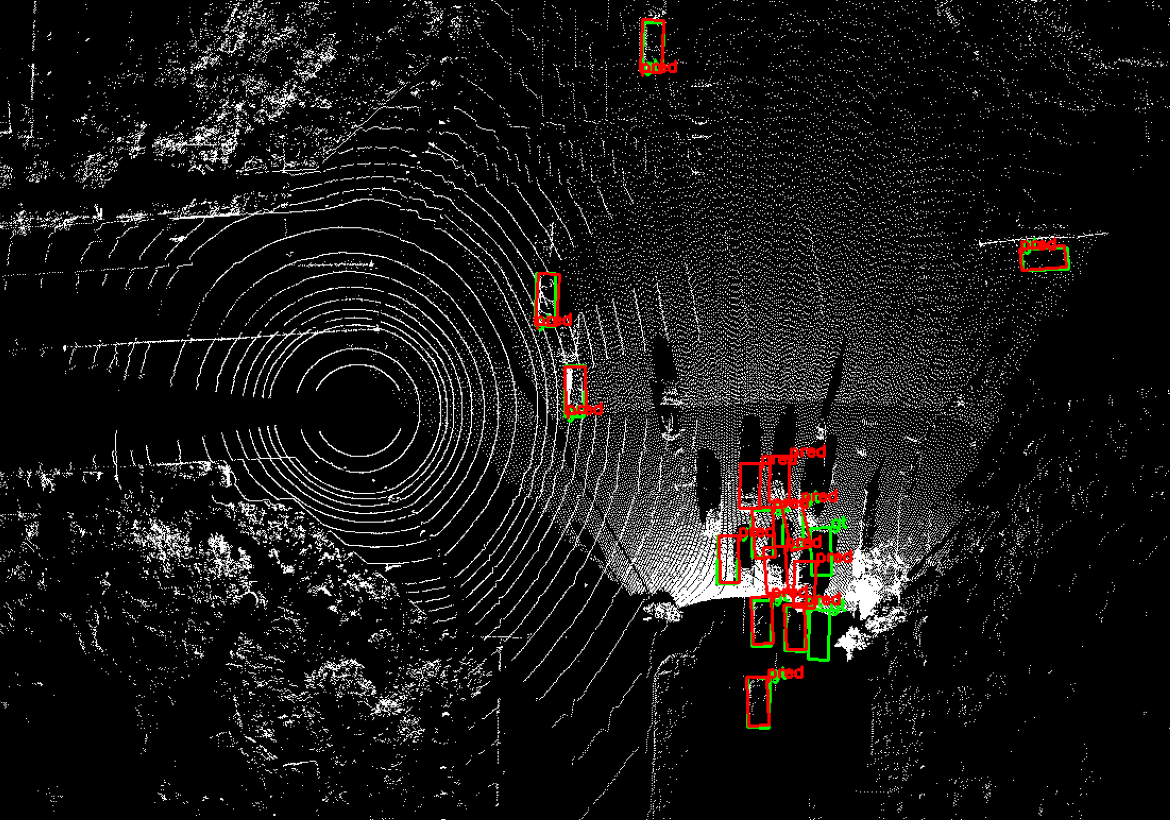}
    \end{minipage}
    }
    \subfigure[Ours]{
    \begin{minipage}[b]{.4\linewidth}
        \centering
        \includegraphics[scale=0.08]{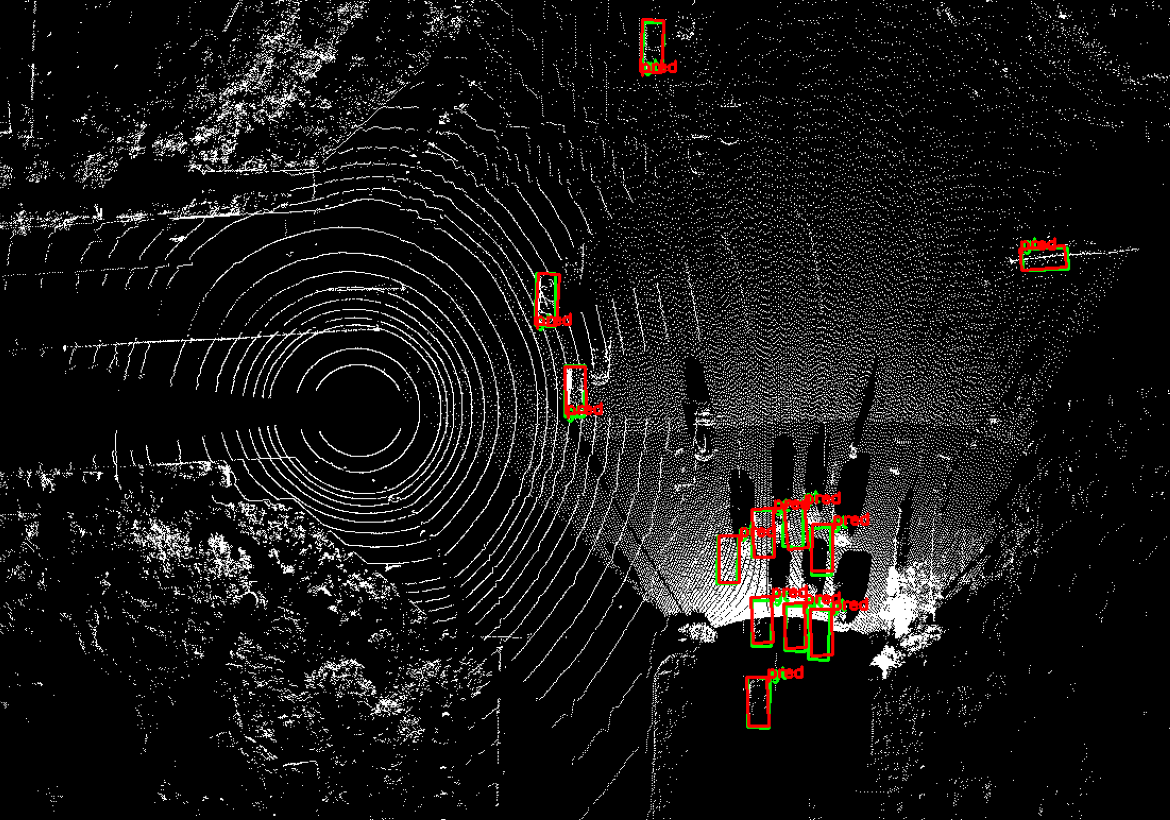}
    \end{minipage}
    }
    \caption{Visualization of perception results on DAIR-V2X dataset. \textcolor{green}{Green} boxes represent ground-truth while \textcolor{red}{red} boxes are prediction. Our method achieves more accurate detection.}
    \label{fig:3}
\end{figure}
\subsection{Qualitative Results}
    Figure \ref{fig:3} shows the detection results under noisy conditions on DAIR-V2X dataset. We can see that FeaKM maintains robustness and predicts accurate bounding boxes even with high pose errors, while other methods suffer from larger displacements. It is noteworthy that due to the dependence of initial pose value, CoAlign exhibits significant degradation when handling large localization errors.

\subsection{Ablation Studies}
    To validate the effectiveness of each module, Table \ref{tab: ablation} shows the AP@0.5 with different noisy levels. We see that the multiscale structure is beneficial for improving the robustness of the model. The reason why foregoing the use of confidence map yields better results at $noise=0$ is that it could provide more interesting points. However, via using confidence map, we can both save computational memory and obtain comparable results.

\section{conclusion}
    This papers proposes a novel collaborative framework FeaKM, which could deal with arbitrary pose errors. We design a keypoints matching strategy to achieve feature-level consistency. FeaKM can maintain robustness and improve the detection performance under seriously noisy conditions. Experiments show that our FeaKM surpasses other SOTA methods. In the future, we aim to achieve more robust collaborative perception based on temporal information.

\begin{acks}
    This work was supported by the National Science and Technology Major Project under Grant 2022ZD0116409.
\end{acks}

\bibliographystyle{ACM-Reference-Format}
\bibliography{reference}

\end{document}